\documentclass[conference]{IEEEtran}
\IEEEoverridecommandlockouts
\usepackage{cite}
\usepackage{amsmath,amssymb,amsfonts}
\usepackage{graphicx}
\usepackage{textcomp}
\usepackage{booktabs}
\usepackage{multirow}
\usepackage{threeparttable}
\usepackage{adjustbox}
\usepackage{placeins}
\usepackage{stfloats}
\usepackage{tabularx}
\usepackage{array}
\usepackage{url}
\usepackage{balance}
\usepackage[hidelinks]{hyperref}
\hypersetup{
	pdftitle={In-Hospital Stroke Risk-State Classification from PPG-Derived Hemodynamic Features},
	pdfauthor={Jiaming Liu, Cheng Ding, Jian Wu, Hongxia Xu, and Daoqiang Zhang},
	pdfkeywords={Clinical time-series mining; Photoplethysmography; In-hospital stroke; Risk-state classification; Temporal anchoring}
}
\graphicspath{{figures/}}
\def\BibTeX{{\rm B\kern-.05em{\sc i\kern-.025em b}\kern-.08em
		T\kern-.1667em\lower.7ex\hbox{E}\kern-.125emX}}

\begin{document}
	
	\title{In-Hospital Stroke Risk-State Classification from PPG-Derived Hemodynamic Features}
	
	\author{
		\IEEEauthorblockN{
			Jiaming Liu\textsuperscript{1},
			Cheng Ding\textsuperscript{1,*},
			Jian Wu\textsuperscript{3,4},
			Hongxia Xu\textsuperscript{3,4},
			Daoqiang Zhang\textsuperscript{1,2}
		}
		\IEEEauthorblockA{
			\textsuperscript{1}\textit{College of Artificial Intelligence, Nanjing University of Aeronautics and Astronautics, China}\\
			\textsuperscript{2}\textit{Key Laboratory of Brain-Machine Intelligence Technology, Ministry of Education, China}\\
			\textsuperscript{3}\textit{State Key Laboratory of Transvascular Implantation Devices and TIDRI, Hangzhou 310009, China}\\
			\textsuperscript{4}\textit{Zhejiang Key Laboratory of Medical Imaging Artificial Intelligence, Hangzhou 310058, China}\\
			\{gaming, chengding, dqzhang\}@nuaa.edu.cn;
			\{wujian2000, Einstein\}@zju.edu.cn\\
			\textsuperscript{*}Corresponding author: Cheng Ding
		}
	}
	
	\maketitle
	
	\begin{abstract}
		The scarcity of temporally aligned pre-event physiological data limits the study of stroke risk states before documented clinical recognition. We focus on patients who experienced stroke during hospitalization while undergoing continuous monitoring, enabling retrospective analysis of pre-anchor photoplethysmography (PPG). Using MIMIC-III and MC-MED, an LLM-assisted pipeline generated candidate stroke anchors from unstructured notes. All retained anchors underwent physician adjudication, and a stratified 100-case double-review audit showed 95.0\% candidate-to-adjudicated agreement within $\pm15$ min. We identified 176 MIMIC-III patients and 154 MC-MED patients with eligible synchronized pre-anchor PPG. A fixed 17-channel hemodynamic representation and ResNet-1D classifier were evaluated using patient-level five-fold internal validation and frozen external testing. At validation-selected operating points, F1-scores were 0.7956, 0.8759, and 0.9406 for 4-, 5-, and 6-hour horizons in MIMIC-III and 0.9256, 0.9595, and 0.9888 in MC-MED; the corresponding threshold-free AUCs ranged from 0.6124 to 0.7079. The PPG model achieved higher F1 than four non-waveform clinical and structured-EHR comparators in all six cohort--horizon settings. On high-risk non-stroke controls, window-level false-positive rates ranged from 0.1371 to 0.2938 and decreased to 0.0183--0.1739 after persistence aggregation. These retrospective findings support measurable pre-anchor PPG structure, but do not establish biological stroke onset, a calibrated bedside alarm, or a clinically validated prediction lead time.
	\end{abstract}
	
	
	\section{Introduction}
	
	\par Stroke is a catastrophic vascular event that demands immediate intervention~\cite{GBD2021StrokeRiskFactorCollaborators2024,Kleindorfer2021}. Yet pre-anchor warning remains difficult to study because patients commonly present after symptom onset. Temporally aligned pre-event physiological data are therefore scarce, leaving the value of continuous monitoring signals such as photoplethysmography (PPG) for hour-scale risk-state characterization unclear~\cite{Hughes2023}.
	
	\par PPG is a plausible sensing modality for this problem because it passively records peripheral blood-volume pulsations and reflects systemic hemodynamic dynamics~\cite{Charlton2022,Elgendi2019}. Cerebrovascular events interact with vascular resistance, blood-flow regulation, and systemic cardiovascular state~\cite{Fan2022}. However, whether pre-anchor PPG morphology contains reproducible information associated with documented in-hospital stroke remains an unvalidated hypothesis~\cite{Alhakeem2025}, largely because datasets rarely capture the transition from an earlier monitored state to a clinically documented event.
	
	\par We address this data limitation by targeting patients who experienced stroke while already hospitalized and continuously monitored. We use the MIMIC-III Clinical Database and matched waveform subset~\cite{Johnson2016MIMICIII,PhysioNet-mimiciii-1.4,PhysioNet-mimic3wdb-matched-1.0} for internal development and MC-MED v1.0.1~\cite{PhysioNet-mc-med-1.0.1,Kansal2025MCMED} for frozen external evaluation. Because structured EHR timestamps often denote documentation rather than the clinically relevant event time, an LLM-assisted pipeline generated candidate anchors from unstructured notes, followed by physician adjudication. This process yielded 176 MIMIC-III and 154 MC-MED patients with eligible synchronized pre-anchor PPG.
	
	\par Hemodynamic features extracted from the pre-anchor waveforms were modeled with ResNet-1D~\cite{He2016,Ribeiro2020,Strodthoff2020}. At validation-selected operating points, F1 increased across 4--6 h horizons, while threshold-free AUC remained more modest. We therefore interpret operating-point performance jointly with class prevalence, frozen external testing, clinical/EHR comparators, and false-alert burden rather than as a calibrated clinical prediction system.
	
	\par
\begin{figure*}[!t]
		\centering
		\includegraphics[width=\textwidth,height=0.6\textheight,keepaspectratio]{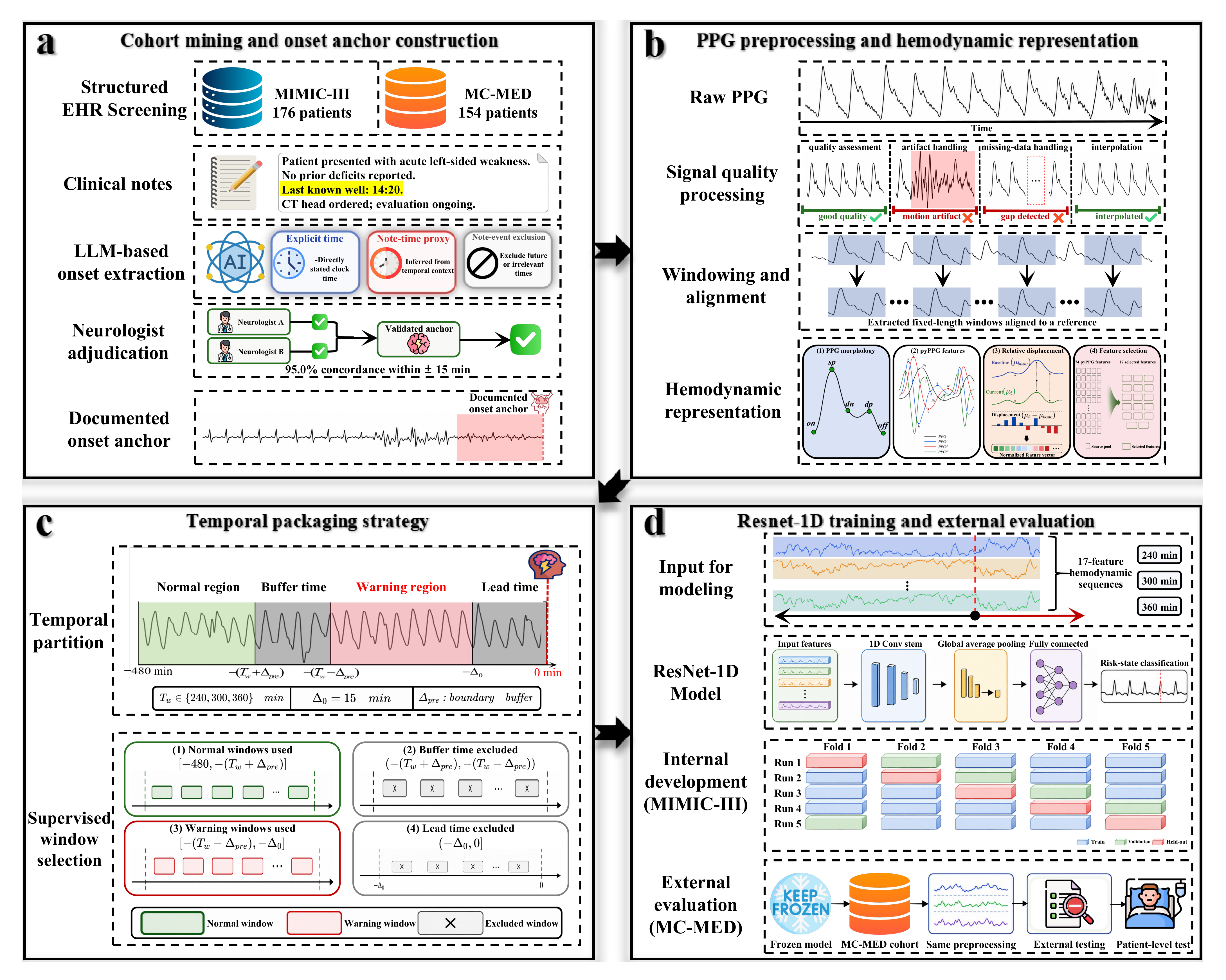}
		\vspace{-4pt}
		\caption{Anchor-aligned PPG mining framework. (a) Cohort construction combines structured EHR screening, note-based onset extraction, and neurologist adjudication to identify documented in-hospital stroke anchors. (b) PPG preprocessing yields quality-controlled, temporally aligned hemodynamic representations. (c) Temporal packaging defines normal, buffer, warning, and lead-time regions around each anchor for supervised window selection. (d) ResNet-1D models are trained with patient-level internal validation on MIMIC-III and evaluated as frozen models on MC-MED.}
		\vspace{-6pt}
		\label{fig:anchor_framework}
	\end{figure*}

 The contribution is data-centric rather than architectural. First, we construct temporally aligned pre-anchor cohorts using LLM-assisted note mining and physician adjudication. Second, we define leakage-aware temporal packaging and a fixed PPG-derived hemodynamic representation using patient-disjoint development and evaluation. Third, we provide patient-level internal validation, frozen cross-source testing, LSTM and non-waveform comparator analyses, feature attribution, subgroup evaluation, anchor perturbation, signal-quality auditing, and high-risk non-stroke false-alert assessment. Fig.~\ref{fig:anchor_framework} summarizes this framework.
	
	\section{Related Work}
	
	\subsection{Stroke Prediction with Electronic Health Records}
	\noindent Stroke stratification has progressed from interpretable clinical scores to machine-learning models based on structured EHR variables~\cite{Dritsas2022,SIRSAT2020}. CHA$_2$DS$_2$-VASc provides long-term thromboembolic risk stratification in atrial-fibrillation populations~\cite{lip2010refining,Kim2017CHA2DS2VASc}, whereas EHR-based studies have used demographics, diagnoses, laboratory tests, vital signs, and medications for stroke prediction~\cite{Lip2022,DBLP2021,SANTOS2022,nwosu2019predicting,Teoh2018StrokeEHR,Yang2021JMIRStroke}. These approaches provide clinically meaningful non-waveform references, but their static or intermittently sampled inputs differ from the within-case, hour-scale waveform task studied here.
	
	\subsection{Physiological Signal Mining for Latent Risk}
	\noindent Deep learning can identify latent disease signatures in physiological signals even when the target event is absent from the recorded segment~\cite{Khurshid2022,Attia2021}. ECG studies have demonstrated near-term atrial-fibrillation risk estimation from arrhythmia-free intervals and other subclinical waveform signatures~\cite{Gadaleta2023,Raghunath2020,Ebinger2023}. These findings motivate temporal risk mining from continuously acquired waveforms, but do not resolve the label-construction problem for acute in-hospital stroke.
	
	\subsection{PPG Hemodynamics and Temporal Labeling}
	\noindent PPG reflects vascular stiffness, aging, pulse morphology, and peripheral blood-volume dynamics~\cite{Bayoumy2021,Charlton2022,Pereira2020,Miller2025,Elgendi2019}. Recent work has inferred stroke-related factors such as cuffless blood pressure and atrial fibrillation from pulse waves~\cite{Mukkamala2022,Lubitz2022,Weng2024}; however, temporally aligned warning of acute in-hospital stroke remains underexplored~\cite{Alhakeem2025}. Existing PPG studies are mainly retrospective or associative~\cite{Li2025,Yang2025} and rarely combine patient-level external testing with deployment-oriented false-alert analysis.
	
	\par The central bottleneck is label availability: clinically relevant stroke times are seldom represented as reliable structured EHR fields, making leakage-aware pre-anchor windows difficult to construct~\cite{Armoundas2024,Sambasivan2021}. We address this bottleneck through LLM-assisted candidate extraction and physician adjudication~\cite{Thirunavukarasu2023}, followed by effective-window auditing, frozen external testing, non-stroke false-alert evaluation, onset-anchor sensitivity analysis, and signal-quality auditing.
	
	\section{Problem Setting}
	\label{sec:problem_setting}
	
	\par This section defines the anchor-aligned, within-case pre-anchor temporal risk-state classification task. We first describe the internal and external clinical waveform resources, then define the temporal classification problem around clinically documented onset anchors, and finally specify the labeling protocol used to construct normal, buffer, warning, and lead-time regions.
	
	\subsection{Data Sources and Cohort Definition}
	\label{sec:data_sources_cohort}
	
	\par We use two large-scale clinical waveform resources: the MIMIC-III Clinical Database and matched waveform subset~\cite{Johnson2016MIMICIII,PhysioNet-mimiciii-1.4,PhysioNet-mimic3wdb-matched-1.0} as the internal development cohort and MC-MED v1.0.1~\cite{PhysioNet-mc-med-1.0.1,Kansal2025MCMED} as the external evaluation cohort.
	
	\par  MIMIC-III (Internal Cohort): Sourced from the Beth Israel Deaconess Medical Center in Boston, MA, this single-center database serves as our primary training and internal validation environment. It comprises comprehensive de-identified health-related data associated with over 40,000 patients who stayed in critical care units between 2001 and 2012. Our analysis focused on the subset of 38,548 patients for whom sufficient waveform data was potentially retrievable. Within this population, the identified stroke subset exhibited a mean age of 66.9 years (SD 15.3), notably higher than the overall cohort mean of 63.8 years. Racial demographics within the stroke subset largely mirrored the general MIMIC population, with 124 (70.5\%) identified as White, 13 (7.4\%) as Black or African American, and 5 (2.8\%) as Asian. Comorbidity profiles indicated significantly higher rates of hypertension in the stroke group (76.7\% vs. 56.8\%) and hyperlipidemia (39.8\% vs. 31.8\%), consistent with established vascular risk factors.
	
	\par MC-MED (External Cohort): MC-MED v1.0.1 is a single-center waveform and EHR resource collected during routine care from adult patients in monitored beds in the Stanford Health Care emergency department between September 2020 and September 2022~\cite{PhysioNet-mc-med-1.0.1,Kansal2025MCMED}. The source database includes 118,385 adult emergency-department visits from 70,545 unique patients. Following the same cohort-construction and waveform-quality criteria, the confirmed stroke cohort included 154 patients and 154 eligible visits. These patients had a mean age of 71.0 years (SD 15.0). The cohort included 74 White patients (48.1\%), 43 Asian patients (27.9\%), and 7 Black or African American patients (4.5\%).
	
	\begin{table*}[t]
		\centering
		\caption{Patient-level characteristics of the source records and retained stroke cohorts}
		\label{tab:cohort_characteristics_mimic_mcmed}
		\scriptsize
		\setlength{\tabcolsep}{4.2pt}
		\renewcommand{\arraystretch}{1.08}
		\begin{tabular*}{\textwidth}{@{\extracolsep{\fill}}llcccc}
			\toprule
			\multirow{2}{*}{Characteristic} & \multirow{2}{*}{Subgroup} &
			\multicolumn{2}{c}{MIMIC-III} &
			\multicolumn{2}{c}{MC-MED} \\
			\cmidrule(lr){3-4}\cmidrule(lr){5-6}
			& & Overall & Stroke subset & Overall & Stroke subset \\
			\midrule
			Sample size & $N$ & 38,548 & 176 & 70,545 & 154 \\
			\midrule
			
			\multirow{2}{*}{Age, years}
			& Mean (SD) & 63.8 (17.5) & 66.9 (15.3) & 51.8 (20.4) & 71.0 (15.0) \\
			& Median [IQR] & 65.7 [52.4--77.9] & 68.8 [54.2--79.6] & 51.0 [34.0--68.0] & 75.0 [61.0--83.0] \\
			\midrule
			
			\multirow{2}{*}{Age group}
			& $<65$ years & 18,769 (48.7\%) & 76 (43.2\%) & 49,038 (69.5\%) & 47 (30.5\%) \\
			& $\geq 65$ years & 19,779 (51.3\%) & 100 (56.8\%) & 21,507 (30.5\%) & 107 (69.5\%) \\
			\midrule
			
			\multirow{4}{*}{Age range}
			& 18--39 years & 4,075 (10.6\%) & 9 (5.1\%) & 23,840 (33.8\%) & 4 (2.6\%) \\
			& 40--59 years & 10,833 (28.1\%) & 52 (29.5\%) & 20,000 (28.4\%) & 30 (19.5\%) \\
			& 60--79 years & 15,775 (40.9\%) & 75 (42.6\%) & 18,829 (26.7\%) & 70 (45.5\%) \\
			& $\geq 80$ years & 7,865 (20.4\%) & 40 (22.7\%) & 7,876 (11.2\%) & 50 (32.5\%) \\
			\midrule
			
			\multirow{3}{*}{Sex}
			& Female & 16,732 (43.4\%) & 88 (50.0\%) & 37,937 (53.8\%) & 74 (48.1\%) \\
			& Male & 21,816 (56.6\%) & 88 (50.0\%) & 32,572 (46.2\%) & 80 (51.9\%) \\
			& Unknown & 0 (0.0\%) & 0 (0.0\%) & 36 (0.1\%) & 0 (0.0\%) \\
			\midrule
			
			\multirow{8}{*}{Race}
			& White & 27,477 (71.3\%) & 124 (70.5\%) & 28,913 (41.0\%) & 74 (48.1\%) \\
			& Asian & 911 (2.4\%) & 5 (2.8\%) & 12,592 (17.8\%) & 43 (27.9\%) \\
			& Black or African American & 2,943 (7.6\%) & 13 (7.4\%) & 3,556 (5.0\%) & 7 (4.5\%) \\
			& Pacific Islander & 11 (0.0\%) & 0 (0.0\%) & 1,197 (1.7\%) & 2 (1.3\%) \\
			& American Indian/Alaska Native & 20 (0.1\%) & 0 (0.0\%) & 169 (0.2\%) & 0 (0.0\%) \\
			& Other & 2,294 (6.0\%) & 21 (11.9\%) & 23,329 (33.1\%) & 28 (18.2\%) \\
			& Declines to state & 0 (0.0\%) & 0 (0.0\%) & 436 (0.6\%) & 0 (0.0\%) \\
			& Unknown & 4,892 (12.7\%) & 13 (7.4\%) & 353 (0.5\%) & 0 (0.0\%) \\
			\midrule
			
			\multirow{5}{*}{Comorbidities}
			& Hypertension & 21,884 (56.8\%) & 135 (76.7\%) & 12,130 (17.2\%) & 57 (37.0\%) \\
			& Diabetes & 10,312 (26.8\%) & 39 (22.2\%) & 6,033 (8.6\%) & 21 (13.6\%) \\
			& Hyperlipidemia & 12,267 (31.8\%) & 70 (39.8\%) & 10,066 (14.3\%) & 41 (26.6\%) \\
			& Chronic kidney disease & 4,896 (12.7\%) & 14 (8.0\%) & 3,562 (5.0\%) & 14 (9.1\%) \\
			& Ischemic heart disease & 13,686 (35.5\%) & 30 (17.0\%) & 4,832 (6.8\%) & 26 (16.9\%) \\
			\bottomrule
		\end{tabular*}
	\end{table*}
	
	\par As summarized in Table~\ref{tab:cohort_characteristics_mimic_mcmed}, both retained stroke subsets skewed older and showed a higher burden of several vascular comorbidities than their respective source records.
	
	\subsection{Anchor-Aligned Stroke Risk-State Classification Task}
	\label{sec:warning_task}
	
	\par We define in-hospital stroke risk-state classification as a within-case, window-level pre-anchor temporal risk-state classification task anchored to the clinically documented stroke time. It is not a stroke-versus-non-stroke clinical screening task. For each retained patient, the note-derived onset anchor is set as $t=0$, and all pre-anchor PPG windows are defined relative to this anchor rather than to the unobserved biological onset. Given $T_w \in \{240,300,360\}$ min, the task is to distinguish warning windows before the anchor from earlier anchor-separated reference windows.
	
	\par Each sample is a packaged PPG-derived hemodynamic window constructed using only information available before the corresponding warning horizon. The anchor time, post-anchor observations, and future windows are never provided as model inputs. To prevent patient leakage, all windows from the same patient are assigned to a single split during internal validation, and the fold-trained models are frozen before external testing on MC-MED.
	
	\subsection{Temporal Labeling Strategy}
	\label{sec:labeling_strategy}
	
	\begin{figure}[t]
		\centering
		\includegraphics[width=\columnwidth]{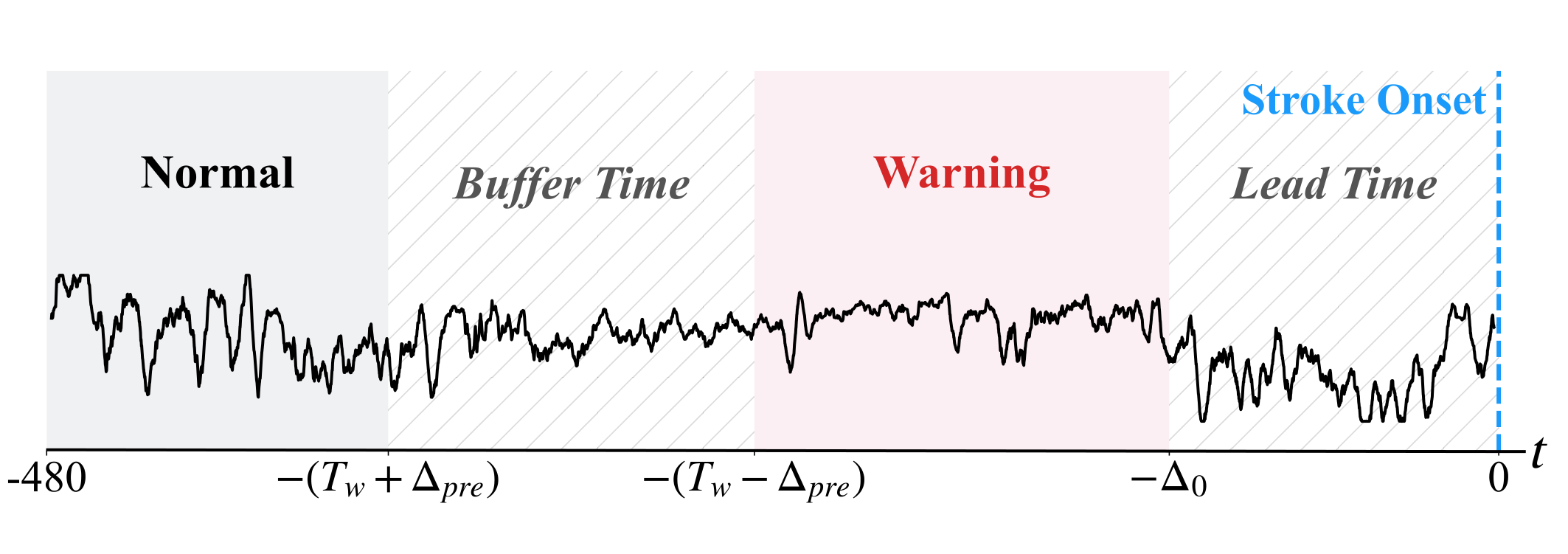}
		\caption{Temporal labeling strategy. Timeline aligned to the clinically documented onset anchor ($t=0$), with exclusion buffers to mitigate label noise and prevent leakage.}
		\label{fig:label_timeline}
	\end{figure}
	
	\par All artifact handling and imputation were restricted within individual source waveform files to avoid creating artificial continuity across unrelated recording sessions.
	
	\par To transform the continuous hemodynamic streams into a supervised learning task, we used a retrospective labeling protocol anchored to the clinically documented onset anchor ($t=0$), as illustrated in Fig.~\ref{fig:label_timeline}. ``Normal'' denotes earlier anchor-separated reference windows, not verified physiological normality. Here, $T_w \in \{240,300,360\}$ min is the warning horizon, $\Delta_{pre}$ is the boundary buffer, and $\Delta_0=15$ min is the lead-time exclusion. The labeling function $y(t)$ is defined as:
	
	\begin{equation}
		y(t) = 
		\begin{cases} 
			1, & \text{if } -(T_{w} - \Delta_{pre}) \le t \le -\Delta_0 \quad (Warning) \\
			0, & \text{if } -480 \le t \le -(T_{w} + \Delta_{pre}) \quad (Normal)
		\end{cases}
	\end{equation}
	
	\par To ensure robust decision boundaries, we implemented a dual-exclusion strategy that discards indeterminate samples falling within the transition zones. As depicted by the hatched regions in Fig.~\ref{fig:label_timeline}, the exclusion set $\mathcal{T}_{\text{drop}}$ is defined as:
	
	\begin{equation}
		\mathcal{T}_{\text{drop}} = \underbrace{[-(T_{w} + \Delta_{pre}), -(T_{w} - \Delta_{pre}))}_{\textbf{Buffer Time}} \cup \underbrace{(-\Delta_0, 0]}_{\textbf{Lead Time}}
	\end{equation}
	
	\noindent The Buffer Time separates earlier reference windows from warning windows, reducing ambiguity during gradual hemodynamic transitions and limiting boundary overfitting~\cite{Hyland2020}. The Lead Time ($\Delta_0 = 15$ min) introduces a short exclusion interval before the documented onset anchor, limiting reliance on near-anchor artifacts.
	
	\section{Proposed Method}
	\label{sec:proposed_method}
	
	\par Following the anchor-aligned task definition, the proposed framework consists of three components: LLM-assisted onset-anchor extraction, PPG hemodynamic feature construction, and a ResNet-1D model for pre-anchor risk-state classification.
	
	\subsection{LLM-Assisted Onset Anchor Extraction}
	\label{sec:llm_onset}
	
	\par Structured EHR fields often lack the temporal resolution required to define pre-recognition windows. We used \texttt{gemini-3-pro-preview} with a fixed, version-controlled prompt to generate candidate temporal anchors from de-identified clinical narratives. Each input contained a row identifier, record timestamp, and clinical text. The extractor prioritized explicitly documented onset, discovery, witnessed, or last-known-well times. If an eligible acute event lacked an explicit time, the record timestamp was retained as a conservative proxy; records without eligible events were assigned \texttt{NULL}.
	
	\par Both cohorts used the same semantic extraction and fallback rules with cohort-specific adapters. For MIMIC-III, inputs came from \texttt{NOTEEVENTS.TEXT} and \texttt{CHARTTIME}. For MC-MED, \texttt{Study} and \texttt{Impression} from \texttt{rads.csv} were concatenated, \texttt{Result\_time} was used as the record timestamp, and \texttt{Order\_time} was retained for audit. Clinical text was used only for anchor generation and never for model training, feature selection, threshold calibration, or prediction.
	
	\par All retained candidate anchors underwent physician adjudication using the same temporal protocol. To quantify temporal agreement without conflating it with inter-rater agreement, a stratified subset of 100 cases was independently reviewed by two neurologists with more than ten years of cerebrovascular experience. In 95.0\% of these cases, the LLM-assisted candidate timestamp was within $\pm15$ min of the final adjudicated anchor. This tolerance matches the near-anchor exclusion interval in Section~\ref{sec:labeling_strategy}; the 95.0\% value therefore denotes candidate-to-adjudicated timestamp agreement, not full-cohort automated-label accuracy.
	
	\par MIMIC-III and MC-MED were accessed under PhysioNet credentialing and applicable data-use agreements. Restricted clinical text, raw LLM outputs, and reviewed patient-level anchors are not redistributed.
	
	\subsection{PPG Hemodynamic Feature Construction}
	\label{sec:hemo_features}
	
	\par Using the adjudicated anchors, we extracted 74 beat-level morphological biomarkers from raw PPG and its derivatives with pyPPG~\cite{goda2023pyppg}. Feature screening used only patient-disjoint MIMIC-III development data: descriptors with negligible class separation (Cohen's $d\leq0.05$) were removed, followed by removal of redundant variables from highly correlated pairs ($|r|>0.80$). MC-MED contributed neither feature selection nor preprocessing estimation.
	
	\par To reduce dependence on absolute amplitude and patient-specific baseline differences, selected descriptors were represented, where applicable, as relative deviations. For a descriptor $x(t)$,
	\begin{equation}
		\mathcal{F}_{\mathrm{rel}}(t)=\frac{x(t)-\mu_{\mathrm{ref}}}{|\mu_{\mathrm{ref}}|+\epsilon},
	\end{equation}
	where $\mu_{\mathrm{ref}}$ is estimated only from that patient's earlier pre-anchor reference period and $\epsilon$ prevents numerical instability near zero. Warning-window outcomes, held-out folds, and external-cohort statistics are excluded from this construction. The final model input contains 17 fixed channels; channel definitions and ordering are documented in the released feature dictionary.
	
	\subsection{Risk-State Classification and Evaluation Strategy}
	\label{sec:modeling_strategy}
	
	\par We use ResNet-1D to model temporal dependencies in PPG-derived hemodynamic sequences. Each input contains 500 consecutive beats represented by 17 channels. The network begins with a 64-channel one-dimensional convolution (kernel size 7, stride 2), batch normalization, rectified linear activation, and max pooling (kernel size 3, stride 2). Four residual stages use $[2,2,2,2]$ basic blocks with $[64,128,256,512]$ channels. Adaptive average pooling and a linear layer produce two class logits. An identically evaluated LSTM is retained as an architectural comparator; the contribution is the temporally aligned data construction and evaluation protocol rather than a new neural architecture.
	
	\par Models are implemented in PyTorch and trained for up to 100 epochs with batch size 128 and random seed 42. Optimization uses Adam with learning rate $10^{-3}$, weight decay $10^{-4}$, a reduce-on-plateau scheduler, and validation-F1 early stopping with patience 15. Class imbalance is addressed with weighted cross-entropy using class weights $[1.0,3.0]$ and a weighted sampler applied only to the training split.
	
	\par Five-fold stratified group cross-validation assigns all windows from one patient to a single fold. Model selection uses only the validation split. Fold-specific operating thresholds are selected from 0.50 to 1.00 in increments of 0.01 by validation $F_2$, followed by $F_1$, recall, AUC, and threshold value as deterministic tie-breakers. This sensitivity-oriented rule explains why threshold-dependent metrics must be interpreted jointly with AUC and false-alert burden. Each selected model and threshold is frozen before MC-MED evaluation; no external feature re-selection, domain adaptation, or threshold recalibration is performed.
	
	\par To assess whether waveform morphology adds information beyond non-waveform context, we also implement an available-component CHA$_2$DS$_2$-VASc baseline~\cite{lip2010refining,Kim2017CHA2DS2VASc} and structured-EHR comparators aligned with Nwosu et al.~\cite{nwosu2019predicting}, Teoh~\cite{Teoh2018StrokeEHR}, and Yang et al.~\cite{Yang2021JMIRStroke}. Comparator variables are restricted to locally available pre-anchor structured data and use the same patient-level folds and frozen external-testing protocol.
	
	\section{Experiments}
	\label{sec:experiments}
	
	\par We first audit the effective analysis units created by temporal anchoring and sequence packaging. We then evaluate ResNet-1D against an LSTM and added clinical/structured-EHR comparators. Finally, we assess feature contributions, subgroup stability, non-stroke alert burden, anchor uncertainty, and signal-quality differences.
	
	\subsection{Experimental Setup}
	\label{sec:experimental_setup}
	
	\par Table~\ref{tab:effective_samples} reports the effective window-level analysis units after temporal anchoring, exclusion-buffer removal, and sequence packaging. Patient-level cohort sizes alone do not characterize the supervision available to the model because training and testing are performed over anchor-aligned waveform windows. Normal denotes earlier anchor-separated windows rather than verified physiological normality.
	
	\begin{table*}[!t]
		\centering
		\caption{Effective window-level analysis units after temporal packaging}
		\label{tab:effective_samples}
		\scriptsize
		\setlength{\tabcolsep}{3.2pt}
		\renewcommand{\arraystretch}{1.08}
		\begin{tabular*}{\textwidth}{@{\extracolsep{\fill}}lrrrrrrrrrrrr}
			\toprule
			\multirow{2}{*}{Horizon} &
			\multicolumn{3}{c}{MIMIC-III train} &
			\multicolumn{3}{c}{MIMIC-III val.} &
			\multicolumn{3}{c}{MIMIC-III test} &
			\multicolumn{3}{c}{MC-MED test} \\
			\cmidrule(lr){2-4}
			\cmidrule(lr){5-7}
			\cmidrule(lr){8-10}
			\cmidrule(lr){11-13}
			& Total & Normal & Warn. 
			& Total & Normal & Warn. 
			& Total & Normal & Warn. 
			& Total & Normal & Warn. \\
			\midrule
			240 min & 20,162 & 6,862 & 13,300 & 2,746 & 891 & 1,855 & 7,205 & 2,420 & 4,785 & 737 & 232 & 505 \\
			300 min & 20,306 & 4,433 & 15,873 & 2,778 & 599 & 2,179 & 7,252 & 1,521 & 5,731 & 748 & 173 & 575 \\
			360 min & 20,510 & 2,349 & 18,161 & 2,778 & 309 & 2,469 & 7,320 & 747 & 6,573 & 752 & 118 & 634 \\
			\bottomrule
		\end{tabular*}
	\end{table*}
	
	\par The observed distribution follows the labeling scheme: longer horizons enlarge warning windows and reduce reference windows, thereby increasing class imbalance. MIMIC-III counts are derived from the internal evaluation partitions, whereas MC-MED counts correspond to the fixed external test set. External metrics are reported from the associated fold-trained frozen models and are interpreted together with AUC and non-stroke false-alert results.
	
	\subsection{Main Risk-State Classification Performance}
	\label{sec:overall_perf}
	
	\par Table~\ref{tab:resnet_lstm_performance} compares ResNet-1D with the identically trained LSTM baseline under patient-level internal evaluation on MIMIC-III and frozen external testing on MC-MED. MIMIC-III results are reported as mean $\pm$ SD over five patient-level held-out folds, whereas MC-MED results are obtained by applying the corresponding fold-trained frozen models without domain adaptation.
	
	\par Because class distributions vary across horizons, no single metric fully characterizes performance. We therefore report accuracy, recall, precision, F1- and F2-scores, and AUC separately, and further examine false-alert burden on non-stroke controls.
	
	\begin{table*}[!t]
		\centering
		\caption{Waveform-model performance across pre-anchor horizons}
		\label{tab:resnet_lstm_performance}
		\scriptsize
		\begin{adjustbox}{width=\textwidth,center}
			\begin{tabular}{c c l c c c c c c}
				\toprule
				Horizon & Cohort & Model & Accuracy & Recall & Precision & F1-score & F2-score & AUC \\
				\midrule
				\multirow{4}{*}{240 min}
				& MIMIC-III & ResNet-1D & $0.6654 \pm 0.0047$ & $0.9833 \pm 0.0106$ & $0.6681 \pm 0.0041$ & $\mathbf{0.7956 \pm 0.0027}$ & $\mathbf{0.8985 \pm 0.0062}$ & $0.6525 \pm 0.0530$ \\
				& MIMIC-III & LSTM      & $0.6645 \pm 0.0061$ & $0.9047 \pm 0.0888$ & $0.6907 \pm 0.0277$ & $0.7801 \pm 0.0161$ & $0.8492 \pm 0.0546$ & $0.6394 \pm 0.0170$ \\
				& MC-MED    & ResNet-1D & $0.8636 \pm 0.0350$ & $0.9341 \pm 0.0413$ & $0.9179 \pm 0.0020$ & $\mathbf{0.9256 \pm 0.0211}$ & $\mathbf{0.9306 \pm 0.0332}$ & $0.6195 \pm 0.0807$ \\
				& MC-MED    & LSTM      & $0.7719 \pm 0.0451$ & $0.8188 \pm 0.0472$ & $0.8361 \pm 0.0399$ & $0.7456 \pm 0.0520$ & $0.7471 \pm 0.0530$ & $0.6129 \pm 0.0491$ \\
				\midrule
				\multirow{4}{*}{300 min}
				& MIMIC-III & ResNet-1D & $0.7860 \pm 0.0129$ & $0.9647 \pm 0.0360$ & $0.8028 \pm 0.0125$ & $\mathbf{0.8759 \pm 0.0105}$ & $\mathbf{0.9269 \pm 0.0243}$ & $0.6124 \pm 0.0668$ \\
				& MIMIC-III & LSTM      & $0.7695 \pm 0.0055$ & $0.9398 \pm 0.0143$ & $0.8015 \pm 0.0039$ & $0.8651 \pm 0.0047$ & $0.9085 \pm 0.0102$ & $0.6649 \pm 0.0115$ \\
				& MC-MED    & ResNet-1D & $0.9229 \pm 0.0273$ & $0.9401 \pm 0.0294$ & $0.9802 \pm 0.0023$ & $\mathbf{0.9595 \pm 0.0151}$ & $\mathbf{0.9478 \pm 0.0238}$ & $0.6847 \pm 0.1560$ \\
				& MC-MED    & LSTM      & $0.9201 \pm 0.0487$ & $0.9367 \pm 0.0545$ & $0.9809 \pm 0.0039$ & $0.9576 \pm 0.0275$ & $0.9447 \pm 0.0441$ & $0.6109 \pm 0.0729$ \\
				\midrule
				\multirow{4}{*}{360 min}
				& MIMIC-III & ResNet-1D & $0.8880 \pm 0.0028$ & $0.9981 \pm 0.0025$ & $0.8894 \pm 0.0013$ & $\mathbf{0.9406 \pm 0.0015}$ & $\mathbf{0.9743 \pm 0.0020}$ & $0.6492 \pm 0.1147$ \\
				& MIMIC-III & LSTM      & $0.8794 \pm 0.0067$ & $0.9697 \pm 0.0123$ & $0.9021 \pm 0.0073$ & $0.9346 \pm 0.0040$ & $0.9553 \pm 0.0085$ & $0.7346 \pm 0.0469$ \\
				& MC-MED    & ResNet-1D & $0.9797 \pm 0.0192$ & $0.9804 \pm 0.0054$ & $0.9975 \pm 0.0008$ & $\mathbf{0.9888 \pm 0.0025}$ & $\mathbf{0.9837 \pm 0.0042}$ & $0.7079 \pm 0.0748$ \\
				& MC-MED    & LSTM      & $0.9715 \pm 0.0171$ & $0.9732 \pm 0.0172$ & $0.9981 \pm 0.0005$ & $0.9854 \pm 0.0088$ & $0.9780 \pm 0.0139$ & $0.7827 \pm 0.0284$ \\
				\bottomrule
			\end{tabular}
		\end{adjustbox}
	\end{table*}
	
	\par ResNet-1D achieved its highest F1-scores at the 360-minute horizon, reaching $0.9406 \pm 0.0015$ on MIMIC-III and $0.9888 \pm 0.0025$ on frozen MC-MED testing. Across the six ResNet-1D settings, AUC ranged from $0.6124$ to $0.7079$. The difference between high operating-point F1 and more modest AUC is expected under the validation-selected, sensitivity-oriented thresholds and horizon-dependent class prevalence: F1 evaluates one operating point, whereas AUC summarizes ranking across thresholds. Both are therefore interpreted together with Table~\ref{tab:effective_samples} and the non-stroke false-alert analysis.
	
	\par The added non-waveform comparators provide clinically relevant reference points rather than stronger waveform architectures. ResNet-1D achieved the highest F1 in all six cohort--horizon settings, but the structured-EHR baselines were occasionally competitive and showed marked external instability in some settings. These results support incremental value from continuous PPG morphology at the selected operating points while reinforcing the need to inspect threshold-free discrimination and transferred-threshold behavior.
	
	\subsection{SHAP-Based Feature Interpretation}
	\label{sec:shap_interpretation}
	
	\par We next examined which PPG-derived features contributed to the model predictions. The analysis was designed to assess cross-dataset consistency and temporal organization, rather than to establish causal biomarkers.
	
	\subsubsection{Cross-dataset feature attribution}
	
\begin{figure}[!t]
		\centering
		\includegraphics[width=\columnwidth]{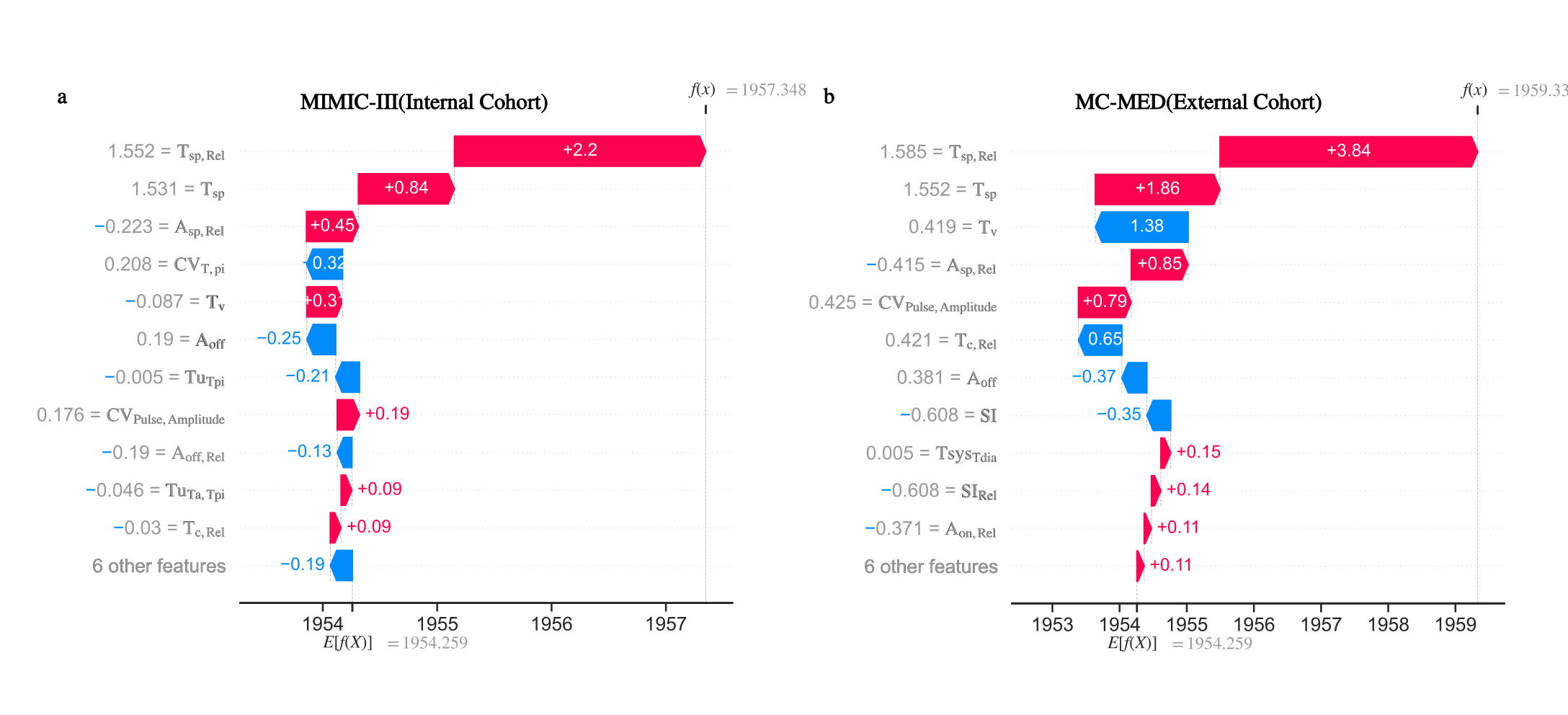}
		\caption{Cross-dataset SHAP analysis. Relative systolic peak time ($T_{sp,Rel}$) is the dominant contributor across MIMIC-III and MC-MED, followed by pulse-timing and amplitude-related features. SHAP values are scaled by $10^3$.}
		\label{fig:shap_waterfall_combined}
	\end{figure}

	We used SHAP~\cite{NIPS2017} to compare feature contributions across cohorts. As shown in Fig.~\ref{fig:shap_waterfall_combined}, relative systolic peak time ($T_{sp,Rel}$) was the leading contributor in both MIMIC-III and MC-MED. Systolic timing, pulse-timing variability, and amplitude-related features also contributed. The cross-cohort recurrence of timing-related descriptors supports an organized morphology signal, but SHAP attribution is descriptive and does not establish stroke-specific or causal biomarkers.
	
	\subsubsection{Temporal consistency of leading features}
	
\begin{figure}[!t]
		\centering
		\includegraphics[width=\columnwidth]{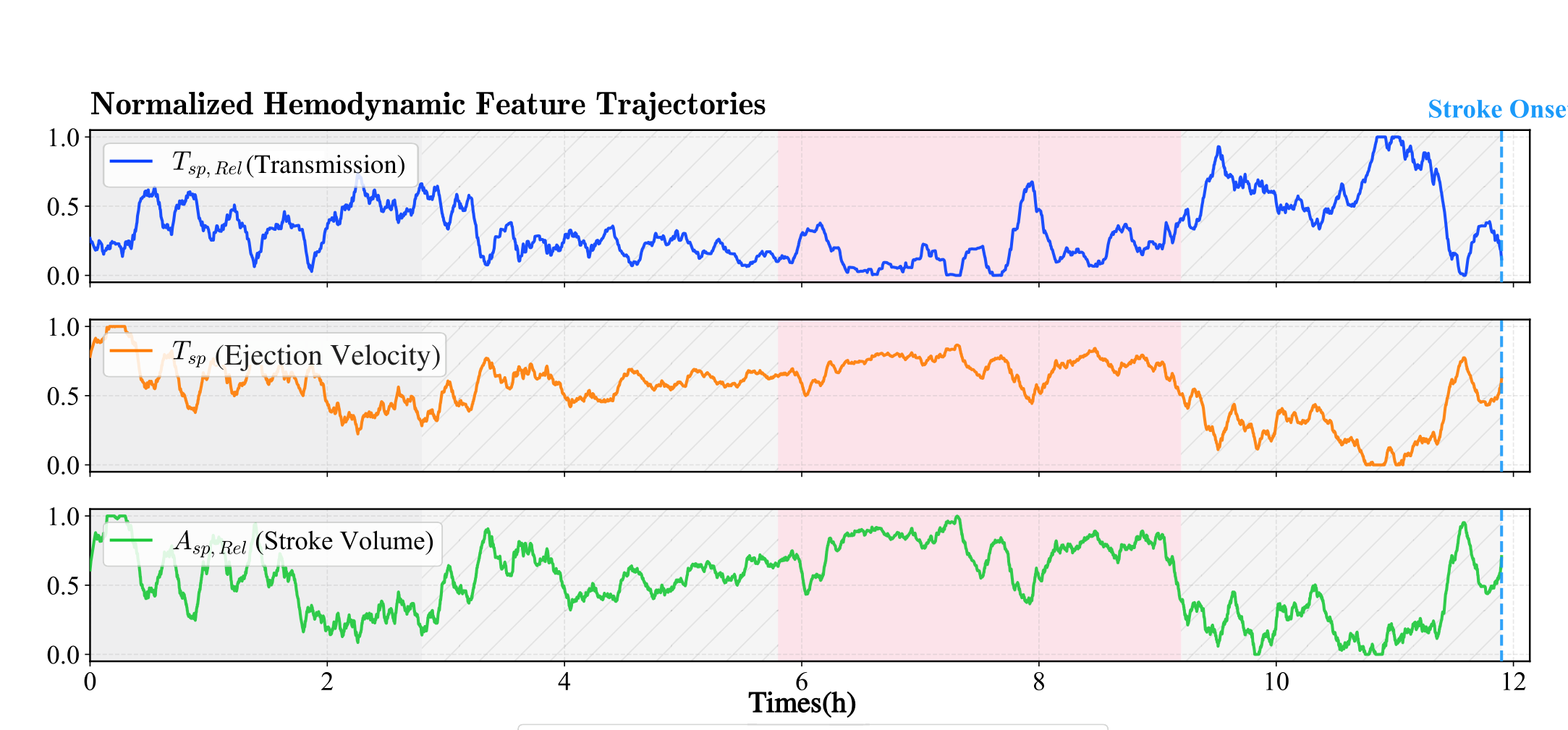}
		\caption{Temporal trajectories of leading contributing factors. In the late pre-anchor interval, relative systolic peak time ($T_{sp,Rel}$) shows an upward drift, while absolute systolic peak time ($T_{sp}$) and relative systolic amplitude ($A_{sp,Rel}$) exhibit downward shifts.}
		\label{fig:feature_trajectory}
	\end{figure}

	Fig.~\ref{fig:feature_trajectory} further examines whether the leading SHAP features changed coherently before the anchor. Relative systolic peak time ($T_{sp,Rel}$) showed a late pre-anchor upward drift. Absolute systolic peak time ($T_{sp}$) and relative systolic amplitude ($A_{sp,Rel}$) showed downward shifts over the same interval. These trajectories provide a descriptive link between feature attribution and pre-anchor hemodynamic dynamics.
	
	\subsection{Subgroup Analysis}
	\label{sec:subgroup_analysis}
	
	\par We examined model performance across clinical risk, race, age, and sex strata in the external MC-MED cohort. Clinical risk was defined using five diagnosis-code-derived comorbidities: hypertension, diabetes mellitus, hyperlipidemia, chronic kidney disease, and ischemic heart disease. Patients were assigned to the low-, medium-, or high-risk group if they had zero, one to two, or at least three comorbidities, respectively. This grouping measures comorbidity burden and is not a validated clinical risk score.
	
	\par
\begin{figure}[!t]
		\centering
		\includegraphics[width=\columnwidth]{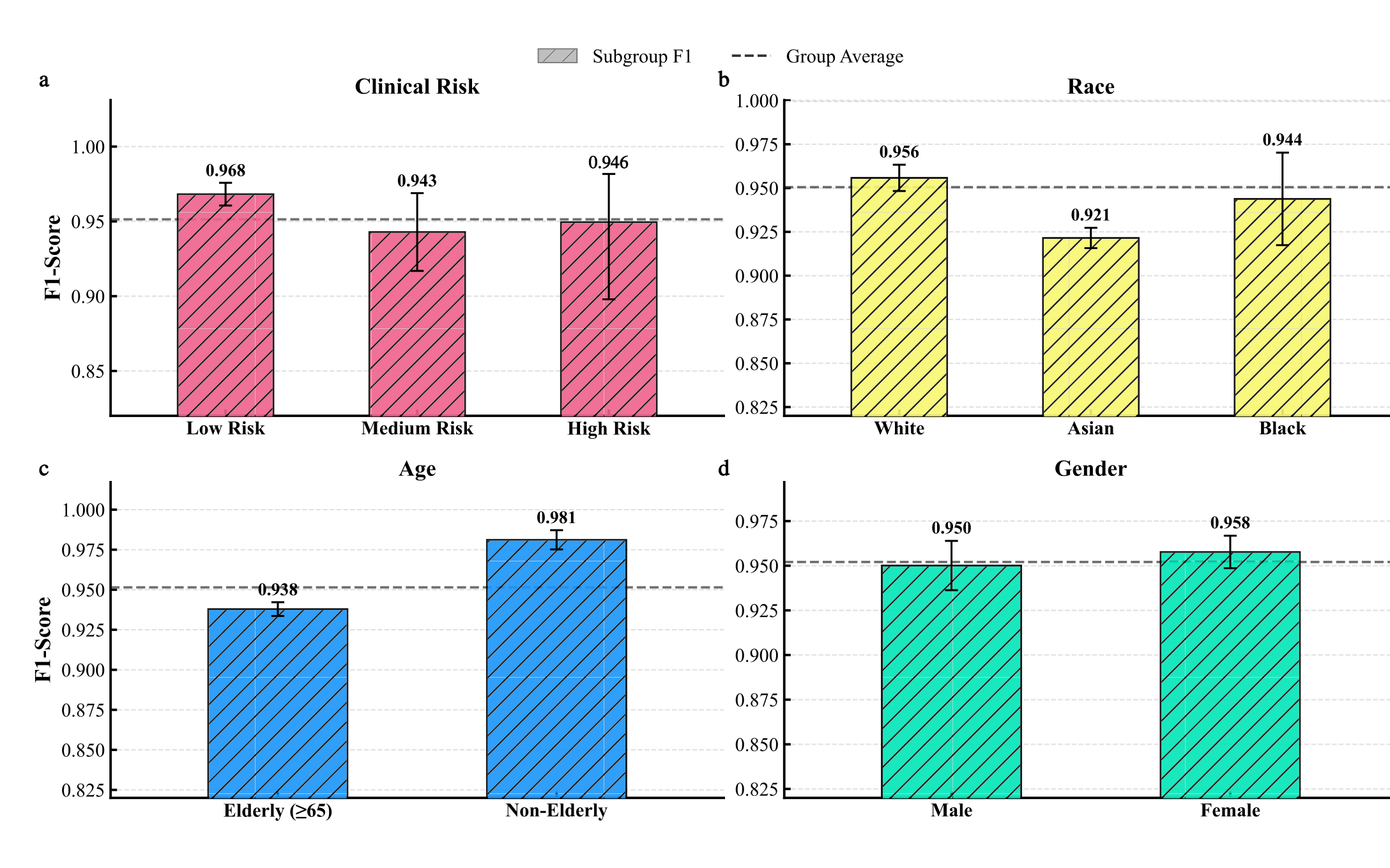}
		\caption{Exploratory subgroup performance on the external MC-MED cohort. Bars denote F1-scores across clinical-risk, race, age, and sex strata, with the dashed line indicating the group-average performance. These analyses are descriptive and characterize subgroup heterogeneity rather than formal demographic fairness or equity.}
		\label{fig:subgroup_analysis}
	\end{figure}

 The frozen model achieved an F1-score above $0.90$ in every subgroup. As shown in Fig.~\ref{fig:subgroup_analysis}, F1-scores were $0.968$, $0.943$, and $0.946$ in the low-, medium-, and high-risk groups, respectively. Across racial groups, F1-scores were $0.956$ for White patients, $0.944$ for Black patients, and $0.921$ for Asian patients. The largest difference was observed between non-elderly and elderly patients ($0.981$ vs. $0.938$). Performance was similar for female and male patients ($0.958$ vs. $0.950$).
	
	\par These results provide a descriptive assessment of subgroup variation. They should not be interpreted as a formal fairness analysis because the evaluation did not include fairness metrics or adjustment for potential confounders.
	
	\begin{table*}[!b]
		\centering
		\caption{Operating-point F1 comparison with added non-waveform clinical and structured-EHR baselines. Values are mean $\pm$ SD over the five fold-trained models; models and thresholds are selected using MIMIC-III development data and frozen for MC-MED.}
		\label{tab:clinical_ehr_baselines}
		\scriptsize
		\setlength{\tabcolsep}{3.6pt}
		\renewcommand{\arraystretch}{1.08}
		
		\begin{tabular*}{\textwidth}{@{\extracolsep{\fill}}lcccccc@{}}
			\toprule
			\multirow{2}{*}{Model} &
			\multicolumn{3}{c}{MIMIC-III F1} &
			\multicolumn{3}{c}{MC-MED F1} \\
			\cmidrule(lr){2-4}\cmidrule(lr){5-7}
			& 240 min & 300 min & 360 min
			& 240 min & 300 min & 360 min \\
			\midrule
			
			ResNet-1D (PPG)
			& $\mathbf{0.7956 \pm 0.0027}$
			& $\mathbf{0.8759 \pm 0.0105}$
			& $\mathbf{0.9406 \pm 0.0015}$
			& $\mathbf{0.9256 \pm 0.0211}$
			& $\mathbf{0.9595 \pm 0.0151}$
			& $\mathbf{0.9888 \pm 0.0025}$ \\
			
			CHA$_2$DS$_2$-VASc$_{\mathrm{avail}}$~\cite{lip2010refining,Kim2017CHA2DS2VASc}
			& $0.6570 \pm 0.0250$
			& $0.7160 \pm 0.0235$
			& $0.7650 \pm 0.0222$
			& $0.8030 \pm 0.0321$
			& $0.8410 \pm 0.0311$
			& $0.8550 \pm 0.0291$ \\
			
			Nwosu-EHR~\cite{nwosu2019predicting}
			& $0.6650 \pm 0.0314$
			& $0.8080 \pm 0.0230$
			& $0.8840 \pm 0.0186$
			& $0.7940 \pm 0.0385$
			& $0.8690 \pm 0.0316$
			& $0.9820 \pm 0.0107$ \\
			
			Teoh-EHR~\cite{Teoh2018StrokeEHR}
			& $0.7260 \pm 0.0268$
			& $0.8060 \pm 0.0212$
			& $0.8800 \pm 0.0196$
			& $0.7440 \pm 0.0441$
			& $0.9340 \pm 0.0288$
			& $0.1300 \pm 0.0464$ \\
			
			Yang-EHR~\cite{Yang2021JMIRStroke}
			& $0.7250 \pm 0.0268$
			& $0.8060 \pm 0.0214$
			& $0.8880 \pm 0.0189$
			& $0.8380 \pm 0.0339$
			& $0.9470 \pm 0.0260$
			& $0.8160 \pm 0.0352$ \\
			
			\bottomrule
		\end{tabular*}
	\end{table*}
	
	\subsection{High-Risk Non-Stroke False-Alert Evaluation}
	\label{sec:nonstroke_false_alert}
	
	\par The primary task compares pre-anchor warning windows with earlier reference windows from patients who later experienced in-hospital stroke. To assess false-alert burden outside this case-only setting, we evaluated the frozen models on separate high-risk non-stroke controls. These controls were older monitored patients with vascular risk factors and no identified stroke diagnosis; they were not intended to represent healthy populations. MIMIC-III controls were aged at least 65 years, had hypertension or diabetes, and remained in the ICU for at least 24 hours. MC-MED controls were aged at least 65 years, had at least one prespecified vascular comorbidity, and had at least 4 hours of usable PPG recordings during an ICU, inpatient, or observation encounter.
	
	\par Both control cohorts were processed using the same 17-feature pipeline and non-overlapping 500-beat sequence packaging procedure as the stroke cohorts. Windows containing any NaN feature value were excluded before inference. For each cohort--horizon pair, the operating threshold was inherited from the corresponding mixed-cohort report and applied unchanged. The control cohorts were not used to select models, thresholds, or aggregation rules.
	
	\par Table~\ref{tab:false_alert_burden} reports packaged windows before filtering ($N_{\mathrm{pkg}}$), NaN-free inference windows ($N_{\mathrm{win}}$), and file-level evaluation identifiers ($N_{\mathrm{ID}}$). ID$+$ denotes the fraction of identifiers with at least five consecutive positive windows. Because each sequence contains 500 beats rather than a fixed number of minutes, temporal aggregation is defined in consecutive windows rather than elapsed time. Evaluation identifiers are file-level packaging groups and may not map one-to-one to unique patients.
	
	\begin{table*}[!t]
		\centering
		\caption{False-alert burden on high-risk non-stroke controls}
		\label{tab:false_alert_burden}
		\footnotesize
		\setlength{\tabcolsep}{5.2pt}
		\renewcommand{\arraystretch}{1.10}
		\begin{tabular*}{\textwidth}{@{\extracolsep{\fill}}c l r r r c c c}
			
			\toprule
			Horizon & Cohort & $N_{\mathrm{pkg}}$ & $N_{\mathrm{win}}$ & $N_{\mathrm{ID}}$ & FPR & Stroke TPR & ID$+$ \\
			\midrule
			240 min & MIMIC-III & 12,972 & 2,997 & 229 & 0.1672 & 0.9833 & 0.0611 \\
			300 min & MIMIC-III & 12,973 & 2,954 & 227 & 0.2881 & 0.9647 & 0.1454 \\
			360 min & MIMIC-III & 12,975 & 2,910 & 230 & 0.2938 & 0.9981 & 0.1739 \\
			\midrule
			240 min & MC-MED    &  7,416 & 1,257 & 165 & 0.1663 & 0.9341 & 0.0303 \\
			300 min & MC-MED    &  7,431 & 1,275 & 159 & 0.1969 & 0.9401 & 0.0377 \\
			360 min & MC-MED    &  7,419 & 1,284 & 164 & 0.1371 & 0.9804 & 0.0183 \\
			\bottomrule
		\end{tabular*}
	\end{table*}
	
	\par Across the corresponding stroke cohorts, TPR remained between $0.9341$ and $0.9981$. On high-risk non-stroke controls, window-level FPR ranged from $0.1672$ to $0.2938$ on MIMIC-III and from $0.1371$ to $0.1969$ on MC-MED. Requiring at least five consecutive positive windows reduced ID$+$ to $0.0611$--$0.1739$ and $0.0183$--$0.0377$, respectively. This reduction shows that temporal persistence can materially suppress isolated alerts, but the remaining burden is not a clinically validated low-alarm regime and could contribute to alarm fatigue. Prospective deployment would require pre-specified persistence rules, cooldown logic, threshold calibration, and direct measurement of clinician workload.
	
	\subsection{Sensitivity Analysis}
	\label{sec:sensitivity_analysis}
	
	\par Finally, we assessed whether the reported performance was sensitive to uncertainty in onset-anchor timestamps or to cohort-specific signal-retention patterns.
	
	\subsubsection{Onset-anchor perturbation}
	\par To assess robustness to timestamp uncertainty, we conducted a post-hoc internal audit on the 240-minute MIMIC-III task. We selected the shortest evaluated horizon because it defines the narrowest warning interval and therefore provides the most stringent test of minute-level anchor uncertainty. For each perturbation, the clinically documented onset anchor was shifted by $\pm 15$, $\pm 30$, or $\pm 60$ minutes. Labels and packaged sequences were then regenerated, and the model was retrained using the same patient-level fold assignments and processing pipeline. Table~\ref{tab:onset_shift_sensitivity} summarizes the results.
	
	\begin{table*}[!t]
		\centering
		\caption{Sensitivity of the 240-minute MIMIC-III task to recognition-anchor perturbation} 
		\label{tab:onset_shift_sensitivity}
		\footnotesize
		\renewcommand{\arraystretch}{1.15}
		\begin{tabular*}{\textwidth}{@{\extracolsep{\fill}}r c c c c c c@{}}
			\toprule
			Anchor shift & Accuracy & Recall & Precision & F1-score & F2-score & AUC \\
			\midrule
			$-60$ min & $0.6150 \pm 0.0168$ & $0.7690 \pm 0.1012$ & $0.6254 \pm 0.0435$ & $0.6845 \pm 0.0274$ & $0.7314 \pm 0.0684$ & $0.6439 \pm 0.0421$ \\
			$-30$ min & $0.6131 \pm 0.0171$ & $0.6677 \pm 0.1128$ & $0.6436 \pm 0.0708$ & $0.6455 \pm 0.0345$ & $0.6567 \pm 0.0822$ & $0.6541 \pm 0.0330$ \\
			$-15$ min & $0.6119 \pm 0.0301$ & $0.6176 \pm 0.0513$ & $0.6395 \pm 0.0224$ & $0.6270 \pm 0.0232$ & $0.6210 \pm 0.0393$ & $0.6468 \pm 0.0436$ \\
			$+15$ min & $0.6135 \pm 0.0186$ & $0.6317 \pm 0.0589$ & $0.6211 \pm 0.0161$ & $0.6253 \pm 0.0318$ & $0.6289 \pm 0.0475$ & $0.6549 \pm 0.0238$ \\
			$+30$ min & $0.6199 \pm 0.0145$ & $0.6303 \pm 0.0742$ & $0.6192 \pm 0.0216$ & $0.6223 \pm 0.0352$ & $0.6265 \pm 0.0583$ & $0.6588 \pm 0.0215$ \\
			$+60$ min & $0.6228 \pm 0.0150$ & $0.6281 \pm 0.0690$ & $0.5958 \pm 0.0269$ & $0.6092 \pm 0.0286$ & $0.6198 \pm 0.0514$ & $0.6654 \pm 0.0196$ \\
			\bottomrule
		\end{tabular*}
	\end{table*}
	
	\par As shown in Table~\ref{tab:onset_shift_sensitivity}, F1-scores ranged from $0.6092 \pm 0.0286$ to $0.6845 \pm 0.0274$, while AUC values ranged from $0.6439 \pm 0.0421$ to $0.6654 \pm 0.0196$. Above-chance discrimination was preserved across all tested shifts. However, the variation in operating-point metrics indicates that precise anchor placement remains relevant when constructing pre-anchor warning labels.
	
	\subsubsection{Signal-quality audit}
	\par We further examined whether external performance could be explained by a systematic recording-quality advantage in MC-MED. Signal quality index (SQI) values are reported as median [IQR]. File retention and duration retention denote the proportions of waveform files and waveform duration retained after quality control, respectively. Cohort-level statistics are reported in Table~\ref{tab:signal_quality_retention}.
	
	\begin{table}[!t]
		\centering
		\caption{Signal quality and waveform retention after quality control}
		\label{tab:signal_quality_retention}
		\footnotesize
		\renewcommand{\arraystretch}{1.12}
		\setlength{\tabcolsep}{8pt}
		\begin{tabular}{@{}l c c c c@{}}
			\toprule
			Cohort & $N$ & SQI & File ret. (\%) & Dur. ret. (\%) \\
			\midrule
			MIMIC-III & 176 & 0.87 [0.79--0.93] & 67.6 & 82.4 \\
			MC-MED    & 154 & 0.93 [0.86--0.96] & 44.6 & 91.2 \\
			\bottomrule
		\end{tabular}
	\end{table}
	
	\par As shown in Table~\ref{tab:signal_quality_retention}, MC-MED had a higher SQI than MIMIC-III ($p < 0.001$) and retained a larger fraction of waveform duration ($91.2\%$ vs. $82.4\%$), whereas MIMIC-III retained more waveform files ($67.6\%$ vs. $44.6\%$). These mixed retention patterns do not support a single uniform external-cohort quality advantage. The higher MC-MED F1 likely reflects a combination of cohort composition, acquisition and documentation differences, retained waveform segments, and transferred operating thresholds; it should not be interpreted as evidence that external evaluation is intrinsically easier or that generalization is complete.
	
	\FloatBarrier
	
	\section{Discussion}
	\label{sec:discussion}
	
	\par The results indicate that routinely acquired PPG contains temporal morphology associated with retrospectively defined pre-anchor warning windows. The strongest contribution is the construction and evaluation of a difficult clinical time-series task: documented anchors are mined from notes, physician-adjudicated, aligned to continuous waveforms, and evaluated without patient overlap or external re-tuning.
	
	\par \textbf{Metric and external-cohort interpretation.} High F1 and modest AUC are not contradictory. Fold-specific thresholds were selected to prioritize $F_2$, and the fraction of warning windows increases with horizon; F1 therefore reflects a sensitivity-oriented operating point and class prevalence, whereas AUC measures threshold-free ranking. MC-MED F1 exceeded internal F1, but AUC and signal-retention differences were mixed. External performance is consequently interpreted as successful transfer under this fixed protocol, not as proof of superior external discriminability.
	
	\par \textbf{Physiological interpretation.} Relative systolic peak time was repeatedly important across cohorts and showed late pre-anchor drift. Systolic peak timing is jointly influenced by ventricular ejection, arterial compliance, vascular tone, and reflected-wave timing~\cite{Charlton2022,Elgendi2019,Fan2022}. The observed attribution and trajectories are therefore physiologically plausible as markers of evolving systemic hemodynamic state, but they are non-specific and cannot establish a causal cerebrovascular mechanism.
	
	\par \textbf{Deployment and limitations.} Window-level false alerts remain non-negligible despite improvement after persistence aggregation, so the present model is not a bedside alarm. The documented anchor may differ from biological onset; the nominal 4--6 h settings are label horizons rather than validated clinical lead times. The independent sample size is determined by 176 and 154 patients, not by the larger number of packaged windows, and small subgroup strata limit precision. Prospective multisite studies must evaluate real-time ingestion, device and site shift, pre-specified thresholds, persistence policies, workflow integration, and alarm fatigue.
	
	\subsection{Reproducibility}
	\par The artifact is available at \url{https://github.com/6jm233333/HemoStroke-PPG-Artifact} and provides cohort scripts, temporal-labeling rules, LLM prompts, the feature dictionary, task registry, baseline-reference outputs, environment files, and commands for regenerating reported statistics and tables. MIMIC-III and MC-MED remain subject to credentialed data-use agreements; restricted records, raw LLM outputs, reviewed anchors, and patient-level derived indices are not redistributed. Experiments used Python~3.11.14 and PyTorch~2.6.0+cu124 on one RTX~3060 GPU with 12~GB memory, an Intel Core i7-14700 CPU, and 32~GB RAM.
	
	\section{Conclusion}
	
	\par We present an anchor-aligned framework for mining pre-anchor PPG hemodynamics in patients who experienced in-hospital stroke. LLM-assisted candidate extraction, physician adjudication, leakage-aware temporal packaging, and patient-level evaluation produced aligned cohorts from MIMIC-III and MC-MED. ResNet-1D achieved F1-scores of $0.7956$--$0.9406$ internally and $0.9256$--$0.9888$ on frozen external testing, while AUC ranged from $0.6124$ to $0.7079$. Added clinical/EHR comparators, false-alert controls, anchor perturbation, signal-quality analysis, subgroup assessment, and feature attribution clarify both the measurable signal and its limits. The findings motivate prospective, calibrated PPG surveillance research; they do not establish biological onset, a validated 4--6 h clinical lead time, or deployment readiness.
	
	\section*{Acknowledgment}
	\par This research was supported by the National Natural Science Foundation of China under Grant No.~32541017 and the Opening Foundation of the State Key Laboratory of Transvascular Implantation Devices under Grant No.~SKLTID2025102. Google Gemini 3 Pro Preview was used only to generate candidate anchors from de-identified clinical notes with a version-controlled prompt. All retained anchors were physician-adjudicated; the system did not determine the final cohort, model labels, analyses, results, or conclusions.
	
	\FloatBarrier
	
	\balance

\end{document}